\documentclass[final]{cvpr}

\usepackage{times}
\usepackage{epsfig}
\usepackage{graphicx}
\usepackage{amsmath}
\usepackage{amssymb}
\usepackage{booktabs}
\usepackage{xcolor}
\usepackage{fancyhdr}
\usepackage{flushend}

\usepackage[pagebackref=true,breaklinks=true,colorlinks,bookmarks=false]{hyperref}

\def\cvprPaperID{16} 
\def\confYear{CVPR 2021}
\newcommand{\tabColor}[1]{\color{darkgray}{#1}\color{black}{}}

\begin{document}

\title{SeqNetVLAD vs PointNetVLAD: Image Sequence vs 3D Point Clouds for Day-Night Place Recognition}

\author{Sourav Garg \qquad Michael Milford \\
QUT Centre for Robotics, Queensland University of Technology\\
{\tt\footnotesize \{s.garg, michael.milford\}@qut.edu.au}
}


\maketitle
\thispagestyle{fancy}
\pagestyle{plain}

\begin{abstract}
Place Recognition is a crucial capability for mobile robot localization and navigation. Image-based or Visual Place Recognition (VPR) is a challenging problem as scene appearance and camera viewpoint can change significantly when places are revisited. Recent VPR methods based on ``sequential representations'' have shown promising results as compared to traditional sequence score aggregation or single image based techniques. In parallel to these endeavors, 3D point clouds based place recognition is also being explored following the advances in deep learning based point cloud processing. However, a key question remains: is an explicit 3D structure based place representation always superior to an implicit ``spatial'' representation based on sequence of RGB images which can inherently learn scene structure. In this extended abstract, we attempt to compare these two types of methods by considering a similar ``metric span'' to represent places. We compare a 3D point cloud based method (PointNetVLAD) with image sequence based methods (SeqNet and others) and showcase that image sequence based techniques approach, and can even surpass, the performance achieved by point cloud based methods for a given metric span. These performance variations can be attributed to differences in data richness of input sensors as well as data accumulation strategies for a mobile robot. While a perfect apple-to-apple comparison may not be feasible for these two different modalities, the presented comparison takes a step in the direction of answering deeper questions regarding spatial representations, relevant to several applications like Autonomous Driving and Augmented/Virtual Reality. Source code available publicly: \url{https://github.com/oravus/seqNet}.
\end{abstract}

\begin{figure}
    \centering
    \includegraphics[width=0.45\textwidth]{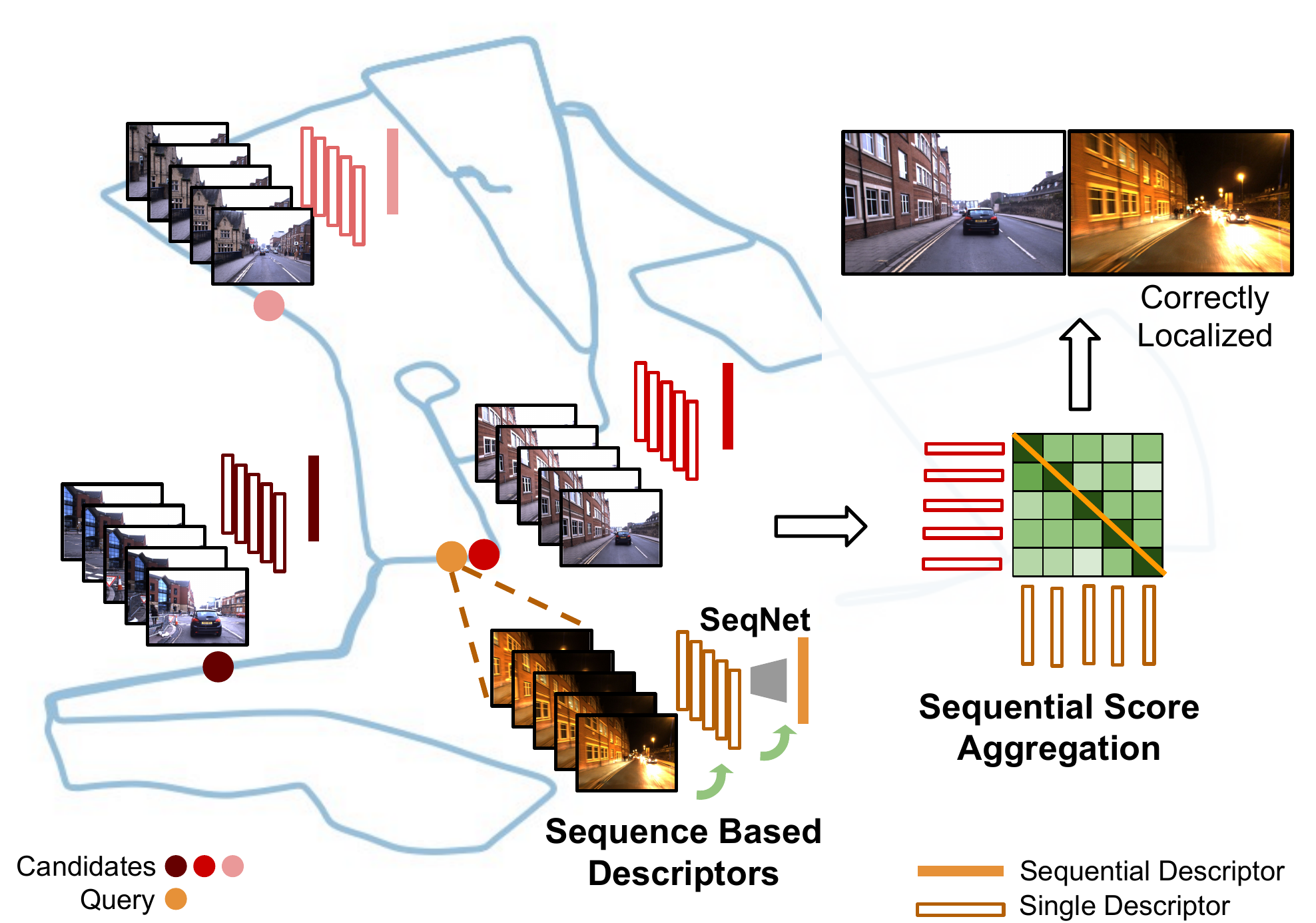}
    \caption{Sequence-based hierarchical visual place recognition. SeqNet learns short sequential descriptors that generate high performance initial match candidates and enables selective control sequence score aggregation using single image learnt descriptors.}
    \label{fig:front-page}
\end{figure}

\section{Introduction}
Visual Place Recognition (VPR) is crucial for mobile robot localization and is typically challenging due to significant changes in scene appearance and camera viewpoint during subsequent visits of known places~\cite{lowry2016visual,gargfischer2021where}. Researchers have explored a variety of methods to deal with this problem ranging from traditional hand-crafted techniques~\cite{cummins2008fab,milford2012seqslam} to modern deep learning-based solutions~\cite{arandjelovic2016netvlad, sarlin2018bleveraging, Lost2018}. Many of these systems aim to push the performance of single image based place recognition by learning better image representations as global descriptors~\cite{arandjelovic2016netvlad, chen2017deep, radenovic2018fine, revaud2019learning} or local descriptors~\cite{detone2018superpoint,dusmanu2019d2,cao2020unifying} and matchers~\cite{sarlin2020superglue}.

To further improve the accuracy of such techniques, researchers have also explored the use of sequential information inherent within the problem of mobile robot localization. However, most of these methods only focus on robustly aggregating \textit{single image} match scores along a sequence~\cite{milford2012seqslam, naseer2014robust, lynen2014placeless, vysotska2015efficient}, where single image representations are agnostic to this post sequential processing. More recent methods have proposed \textit{sequential descriptors}~\cite{garg2020delta, facil2019condition, neubert2019neurologically, arroyo2015towards, chancan2020deepseqslam} that generate place representations considering sequential imagery \textit{before} any sequence score aggregation. \cite{garg2021seqnet} proposed SeqNet and a hierarchical VPR pipeline where sequential descriptors are used to select match hypotheses for single image based sequence score aggregation, as shown in Figure~\ref{fig:front-page}. 

A parallel line of research for place recognition exists with regards to using 3D data in the form of point clouds, as done in PointNetVLAD~\cite{angelina2018pointnetvlad}, DH3D~\cite{du2020dh3d} and others~\cite{liu2019lpd,hui2021efficient,sun2020dagc,vid2021locus,kim2018scan}. Instead of using single images or image sequences, these methods rely on point cloud data, typically captured through a LiDAR sensor. Using 3D information in this fashion has significant advantages when considering extreme appearance variations, for example, matching data across day vs night, where single image-based solutions fail catastrophically. \cite{angelina2018pointnetvlad} demonstrates this behavior by comparing their method against NetVLAD~\cite{arandjelovic2016netvlad}. However, it is not known how well image-based methods compare against a 3D point cloud based technique when a sequence of images is considered. In this extended abstract, we conduct additional experiments with SeqNet~\cite{garg2021seqnet}, showcasing that image sequences can potentially outperform 3D point cloud based methods given a similar metric span and localization radius. As a preliminary investigation's late-breaking result, we only consider a single dataset for this analysis: Oxford Robotcar~\cite{maddern20171}, which was originally used by both SeqNet~\cite{garg2021seqnet} and PointNetVLAD~\cite{angelina2018pointnetvlad}.

We refer the readers to the original works~\cite{angelina2018pointnetvlad,garg2021seqnet} for detailed methodology description. Here, we primarily focus on the experimental settings and results.

\begin{figure}
    \centering
    \includegraphics[scale=0.3]{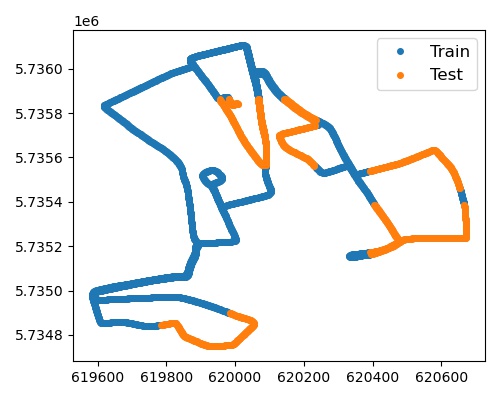}
    \includegraphics[scale=0.3]{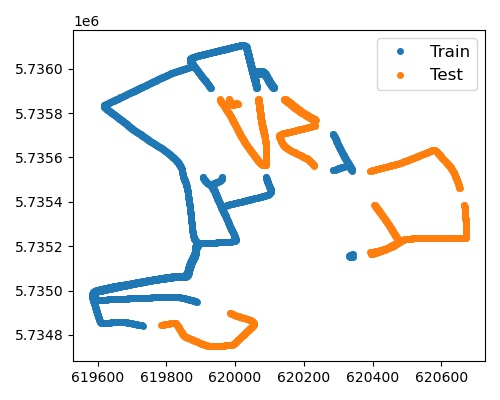}
    \caption{(Left) Train and test splits for the Oxford Robotcar dataset as used in~\cite{angelina2018pointnetvlad} and (right) an updated train split instance ensuring no visual overlap between train and test imagery.}
    \label{fig:splits}
\end{figure}

\setlength{\topmargin}{-24pt}
\setlength{\headheight}{0pt}
\section{Experimental Settings}

\subsection{Dataset}
We use two traverses from the Oxford Robotcar dataset: one from day time (2015-03-17-11-08-44) and the other from night time (2014-12-16-18-44-24). Both these traverses were used in the original works, however, train and test splits differed. Hence, we use the splits defined by PointNetVLAD\footnote{https://github.com/mikacuy/pointnetvlad} to train and test SeqNet\footnote{https://github.com/oravus/seqNet}. Note that the training and test splits are captured from geographically disparate locations and each split comprises its own reference (day) and query (night) database.

The LiDAR based 3D point cloud data in the Oxford dataset used for benchmarking PointNetVLAD~\cite{angelina2018pointnetvlad} strictly captures the local surroundings. On the other hand, forward-facing RGB images can comprise information projected from locations much farther away from the camera. Thus, the train-test split defined in~\cite{angelina2018pointnetvlad} (see Figure~\ref{fig:splits} (left)) potentially leads to visual overlap between both the splits when using image data. Therefore, keeping the test split the same, we additionally created an updated instance of the train split avoiding such visual overlap between the two splits, as shown in Figure~\ref{fig:splits} (right). In Table~\ref{tab:perfComp}, * marked results presented in \tabColor{gray} font color correspond to the usage of the revised train split. For both the split settings, the best (bold) and the second best (italics) results are formatted independently in Table~\ref{tab:perfComp}.

\subsection{Metric Span}
For PointNetVLAD, we use the authors' provided implementation for extracting point clouds and corresponding place representations, which have a metric span of $20$ meters per place. For SeqNet and other sequence based methods, we use an image sequence of length $5$ with a fixed frame separation of $2$ meters between adjacent frames, leading to a metric span of $10$ meters per place\footnote{Since SeqNet results with a metric span of $10$ meters were found to be superior to PointNetVLAD, we did not conduct further experiments with a larger metric span for SeqNet.}. 

\begin{table}
\caption{Performance Comparison - Oxford (Day vs Night): Recall@K (1,5,20)}
    \centering
    \begin{tabular}{l ll}
        \toprule
        \textbf{Method}  & Oxford Robotcar \\
        \midrule

\textbf{Single Image Descriptors:} & \\
 
 
        NetVLAD~\cite{arandjelovic2016netvlad}  & 
 0.54/0.74/0.89 \\

        NetVLAD-FT ($S_1$~\cite{garg2021seqnet})  & 
        0.62/0.83/0.94 \\
        \tabColor{NetVLAD-FT* ($S_1$~\cite{garg2021seqnet})}  & 
        \tabColor{0.59}/\tabColor{0.78}/\tabColor{0.92} \\
\midrule  
\textbf{Point Clouds Descriptors:} & \\
        PointNetVLAD (Base)~\cite{angelina2018pointnetvlad} &
        0.77/0.92/0.96 \\
        PointNetVLAD (Refine)~\cite{angelina2018pointnetvlad} &
        0.76/0.90/0.94 \\       

\midrule      
\textbf{Sequential Descriptors:} & \\
        SmoothNetVLAD~\cite{garg2020delta}   & 
        0.66/0.75/0.87 \\

        DeltaNetVLAD~\cite{garg2020delta}  & 
        0.41/0.64/0.84 \\

        SeqNetVLAD ($S_5$)  &  
        \textit{0.87}/\textit{0.94}/\textbf{0.99} \\
        \tabColor{SeqNetVLAD* ($S_5$)} &  
        \textbf{\tabColor{0.85}}/\textit{\tabColor{0.91}}/\textbf{\tabColor{0.98}} \\
\midrule  
\textbf{Sequential Score Aggregation:} & \\
        SeqMatch-NetVLAD~\cite{milford2012seqslam}   & 
        0.67/0.78/0.89 \\ 

         SeqMatch-NetVLAD-FT~\cite{milford2012seqslam}   & 
      0.84/0.92/0.98 \\
         \tabColor{SeqMatch-NetVLAD-FT*}~\cite{milford2012seqslam}   & 
      \tabColor{0.79}/\tabColor{0.88}/\tabColor{0.96} \\


        SeqNetVLAD-HVPR ($S_5$ to $S_1$) & 
      \textbf{0.88}/\textbf{0.96}/\textbf{0.99} \\
        \tabColor{SeqNetVLAD-HVPR* ($S_5$ to $S_1$)} & 
      \textit{\tabColor{0.83}}/\textbf{\tabColor{0.93}}/\textbf{\tabColor{0.98}} \\
        
        \bottomrule
    \end{tabular}
    \label{tab:perfComp}
\end{table}

\subsection{Descriptors Details} 
PointNetVLAD descriptors are of size $256$ as the authors observed no notable performance gain with further doubling of descriptor dimensions~\cite{angelina2018pointnetvlad}. For SeqNet, 4096-dimensional PCA'd NetVLAD descriptors were used as the underlying single image representations and output was a 4096-dimensional sequential descriptor, as per the original setting~\cite{garg2021seqnet}. From here on, we refer to this SeqNet descriptor based on NetVLAD as \textit{SeqNetVLAD}.

\subsection{Evaluation}
We use Recall@$K$ ($K\in \{1,5,20\}$) as the evaluation metric as also used by both PointNetVLAD and SeqNet. Localization radius is set to be 25 meters as used by PointNetVLAD.

\section{Results}
Table~\ref{tab:perfComp} shows performance comparison between different types of approaches to place recognition: \textbf{1)} Single Image Descriptors including NetVLAD~\cite{arandjelovic2016netvlad} and its nighttime fine-tuned version NetVLAD-FT (trained as $S_1$, as described in~\cite{garg2021seqnet}); \textbf{2)} Point Cloud Descriptors including PointNetVLAD with both its \textit{base} version trained only on the Oxford Robotcar dataset and \textit{refine} version trained on multiple datasets; \textbf{3)} Sequential Descriptors including Smoothed and Delta Descriptor defined using NetVLAD, as described in~\cite{garg2020delta} and SeqNetVLAD~\cite{garg2021seqnet}; and \textbf{4)} Sequential Score Aggregation including SeqSLAM-based~\cite{milford2012seqslam} sequence matching defined on single image descriptors using NetVLAD and NetVLAD-FT, referred to as SeqMatch-NetVLAD and SeqMatch-NetVLAD-FT respectively, and a hierarchical approach as per~\cite{garg2021seqnet}, referred to as SeqNetVLAD-HVPR, where SeqNetVLAD is used as a sequential descriptor to select top matching candidates for SeqMatch-NetVLAD-FT.

It can be observed from Table~\ref{tab:perfComp} that sequence-based methods like SmoothNetVLAD ($0.66$), SeqMatch-NetVLAD ($0.67$) improve performance on top of single image only techniques NetVLAD ($0.54$) and NetVLAD-FT (\tabColor{$0.59$}/$0.62$) but do not approach performance of PointNetVLAD ($0.77$). However, SeqMatch-NetVLAD-FT (\tabColor{$0.79$}/$0.84$), SeqNetVLAD (\tabColor{$0.83$}/$0.87$) and SeqNetVLAD-HVPR (\tabColor{$0.85$}/$0.88$) surpass PointNetVLAD's performance. This demonstrates that not only the sequential information is a strong cue for place recognition under challenging appearance conditions, trained sequential descriptors~\cite{garg2021seqnet} might be learning the underlying 3D scene structure implicitly, leading to better performance than what was achievable through traditional sequence-based methods~\cite{milford2012seqslam}.

The experiments conducted in this preliminary investigation have their limitations as a perfect apple-to-apple comparison between RGB image sequence and LiDAR point clouds is not trivial. Both the sensor modalities have complementary characteristics. As compared to RGB cameras, active range sensors like LiDARs are not drastically affected by variations in environmental conditions such as time of day and seasonal cycles. However, the inherent information richness of RGB image sensors, compared to LiDAR point clouds which are  typically sparse, makes room for advanced image processing techniques, potentially leading to improved performance even under challenging environmental conditions. Furthermore, as a robot moves through an environment, the data accumulation strategy also plays a key role in determining the robustness of a place representation. For example, the strategy of feeding single images to sequence score aggregation methods~\cite{milford2012seqslam,naseer2014robust,lynen2014placeless} or sequential descriptor networks~\cite{garg2021seqnet,facil2019condition,chancan2020deepseqslam} can also be emulated for point cloud based techniques that currently \textit{pre-process} individual point clouds to form a relatively larger one~\cite{angelina2018pointnetvlad,du2020dh3d} before learning any place representations.

\section{Conclusion}
With recent advances in deep learning, several novel methods have been developed for place (spatial) representations including both 3D point cloud based~\cite{angelina2018pointnetvlad,du2020dh3d} and those based on image sequences~\cite{garg2021seqnet,facil2019condition,garg2020delta}. Both these modalities have their own inherent characteristics and there remain several questions unanswered in terms of what might constitute an ideal representation of the world perceived by a mobile robot~\cite{davison2018futuremapping, garg2020semantics}. The analysis presented in this extended abstract takes an initial step towards answering such questions with preliminary investigations. Future work will investigate the scope of combining image sequences and 3D information for an even further improved spatial understanding as also explored recently in~\cite{garg2019look, oertel2020augmenting}.


\bibliographystyle{ieee_fullname}
\bibliography{Region_NetVLAD,sg,egbib,moreRefs}

\end{document}